\documentclass[final,5p,twocolumn,authoryear]{elsarticle}

\usepackage{amssymb}
\usepackage{verbatim}
\usepackage{amsmath}

\usepackage{latexsym}
\usepackage{threeparttable}
\usepackage{amssymb}
\usepackage{amsmath}
\usepackage{amsthm}
\usepackage{booktabs}
\usepackage{graphicx}
\usepackage{color}

\usepackage{subcaption}
\usepackage{arydshln}

\usepackage{algorithm}
\usepackage{algorithmic}
\usepackage{array}
\usepackage{multirow}
\usepackage{xspace}

\usepackage[utf8]{inputenc}
\usepackage{colortbl}
\usepackage{xcolor}
\usepackage[normalem]{ulem}

\newcommand{\Method}{ms-Mamba\xspace}

\newcommand{\first}[1]{\textbf{{#1}}}
\newcommand{\second}[1]{\underline{{{#1}}}}
\newcommand{\etal}{\textit{et al.}\xspace}

\newcommand{\new}[1]{{#1}}
\newcommand{\del}[1]{} 

\journal{Neurocomputing}

\begin{document}

\begin{frontmatter}


\title{\Method: Multi-scale Mamba for Time-Series Forecasting\tnoteref{label1}}
\author[ceng]{Yusuf Meric Karadag\corref{cor1}}
\ead{meric.karadag@metu.edu.tr}
\affiliation[ceng]{organization={Dept. of Computer Eng. and ROMER Robotics Center, Middle East Technical University},
            state={Ankara},
            country={Turkey}}

\affiliation[arch]{organization={Dept. of Architecture and ROMER Robotics Center, Middle East Technical University},
            state={Ankara},
            country={Turkey}}


\author[ceng]{Ismail Talaz}
\author[arch]{Ipek Gursel Dino}
\author[ceng]{Sinan Kalkan}

\begin{abstract}
The problem of Time-series Forecasting is generally addressed by recurrent, Transformer-based and the recently proposed Mamba-based architectures. However, existing architectures generally process their input at a single temporal scale, which may be sub-optimal for many tasks where information changes over multiple time scales. In this paper, we introduce a novel architecture called Multi-scale Mamba (\Method) to address this gap. \Method incorporates multiple temporal scales by using multiple Mamba blocks with different sampling rates ($\Delta$s). Our experiments on many benchmarks demonstrate that \Method outperforms state-of-the-art approaches, including the recently proposed Transformer-based and Mamba-based models. \new{For example, on the Solar-Energy dataset, \Method outperforms its closest competitor S-Mamba (0.229 vs. 0.240 in terms of mean-squared error) while using fewer parameters (3.53M vs. 4.77M), less memory (13.46MB vs. 18.18MB), and less operations (14.93G vs. 20.53G MACs), averaged across four forecast lengths.} Codes and models will be made available. 
\end{abstract}






\end{frontmatter}



\section{Introduction}

Time-series Forecasting (TSF) is the problem of predicting future values of a variable of interest, given its history \citep{lim2021time,miller2024survey,nobrega2019sequential,george2023online}. This fundamental problem used to be generally addressed using recurrent architectures \citep{williams1989learning,elman1990finding} and long-short term memory \citep{hochreiter1997long} or their variants \citep{chung2014empirical,graves2005framewise}, see, e.g., \citep{lim2021time,miller2024survey} for detailed surveys. Such models are inherently well-suited to the task due to their sequential information modeling abilities. The introduction of self-attention based architectures, a.k.a. Transformers \citep{vaswani2017attention}, enabled attending to more informative patterns and correlations across time and provided significant improvements. However, Transformers' quadratic computational complexity has been a limiting factor.

State Space Models (SSMs) \citep{gu2021efficiently,smith2022simplified} are reported to provide a better balance between performance and computational complexity. Recently proposed architectures based on SSMs, namely, Mamba \citep{gu2023mamba}, offer the promise of on-par or better performance than Transformer-based alternatives while running significantly faster. This has led to the wide-spread use of Mamba or its derivatives across different domains \citep{gu2023mamba,yue2024medmamba,qu2024survey,xu2024survey}.

The use of a Mamba-based approach for TSF was recently explored by \cite{wang2024mamba}. Wang \etal introduced an architecture, called S-Mamba, which used Mamba in both the forward and reverse directions for TSF. Wang \etal showed that this simple approach obtained state-of-the-art (SOTA) results on many TSF benchmarks, often providing significant gains over Transformer-based architectures.

\begin{figure}[hbt!]
    \centering
    \includegraphics[width=0.99\linewidth]{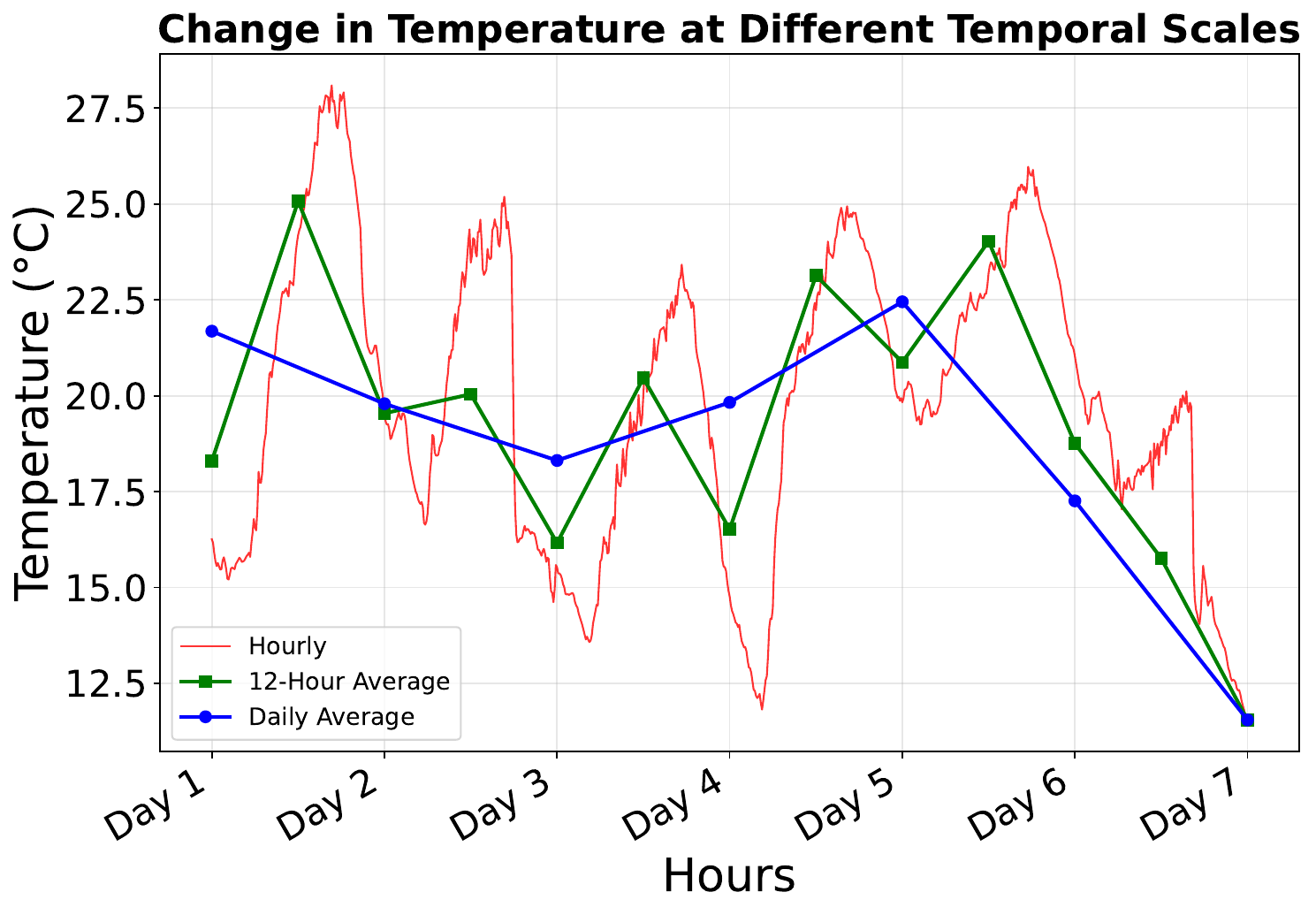}
     \includegraphics[width=0.99\linewidth]{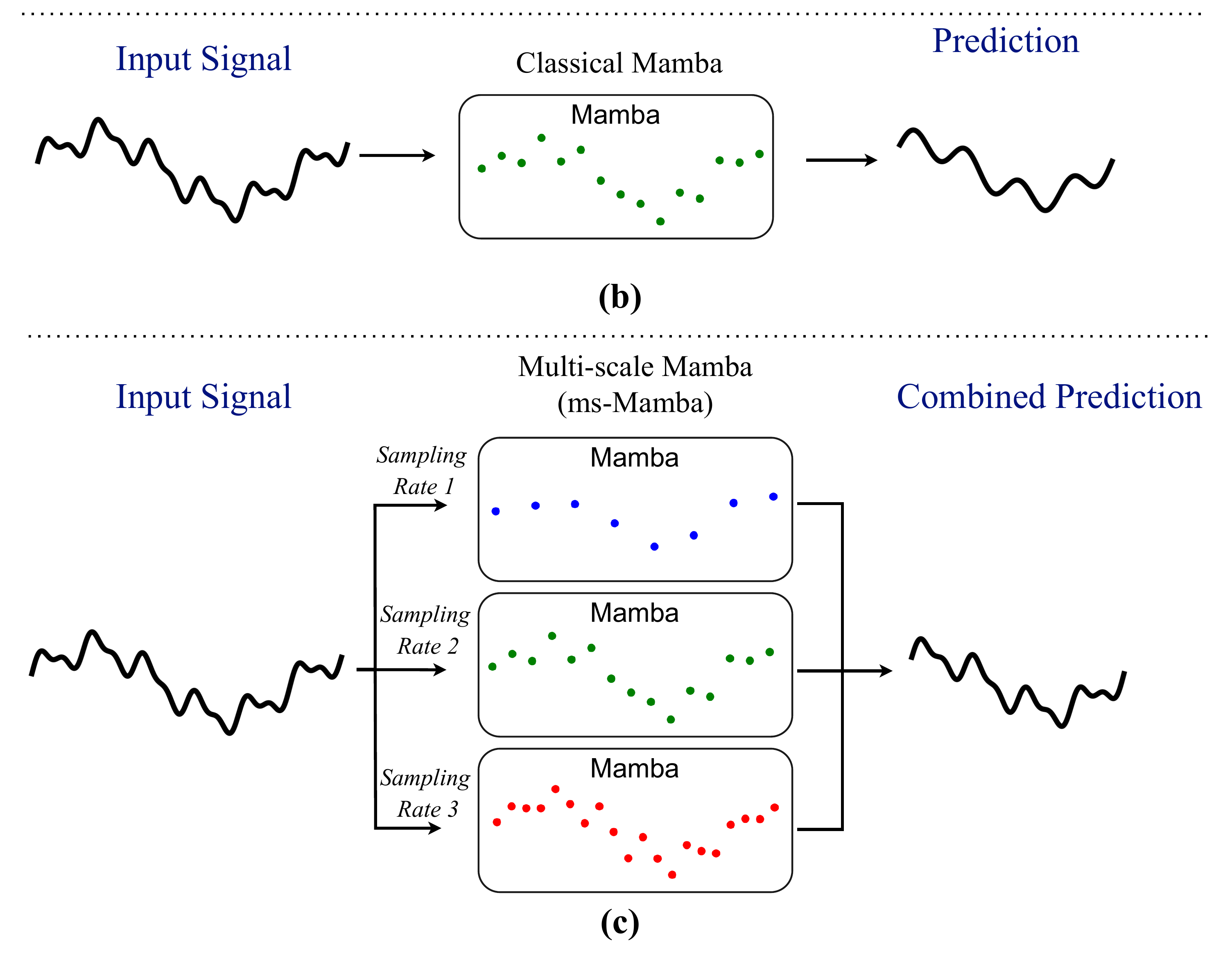}
    \caption{
    \textbf{(a)} Time-series data often contain information at multiple time-scales. Illustrated here is temperature from the Weather dataset \cite{wu2021autoformer}, averaged over different window lengths to highlight different scales. \new{See \ref{sec:appendix_vis} for similar multi-scale visualizations from other datasets.} \textbf{(b)} Mamba and its variations (S-Mamba) use a single time-scale while processing time-series data. \textbf{(c)} Our \Method processes its input at different time-scales to better capture signal at different scales.}
    \label{fig:teaser}
    \vspace*{0.55cm}
\end{figure}


\noindent\textbf{Motivation: Why multi-scale TSF?} Time-series data generally consist of signals of multiple temporal scales (see, e.g., Figure \ref{fig:teaser}(a)). A temperature signal, for example, can have trends at scales of hours, days (day and night), weeks, months or years. To better capture and exploit the multi-scale nature of time-series data, the literature has introduced extensions over the conventional models; e.g., multi-scale recurrent architectures \citep{chung2017hierarchical}, multi-scale convolution \citep{li2024multi} and multi-scale Transformers \citep{zhang2024multi,chen2024pathformer}. 

\noindent\textbf{Our Solution}. In this paper, we introduce \Method, a Mamba-based architecture for multi-scale processing of time-series data (Figure \ref{fig:teaser}(c)), \new{in contrast to standard Mamba-based models that process data at a single time-scale (Figure \ref{fig:teaser}(b))}. To be specific, by leveraging on the versatility of SSMs' learnable sampling rate, we construct a block that consists of multiple SSMs with different independent or inter-related sampling rates. We show that our \Method performs better than Transformer-based and Mamba-based architectures on several datasets.

\noindent\textbf{Contributions}. Our main contributions are as follows:
\begin{itemize}
    \item We propose a multi-scale architecture based on Mamba. To do so, we use multiple SSMs with different sampling rates to process the signal at different temporal scales.

    \item We introduce and compare different strategies for using different sampling rates for different SSMs: (1) Using hyper-parameters as multipliers for a learned sampling rate, (2) learning different sampling rate for each SSM,  (3) and estimating sampling rates from the input. 

    \item We show that, on the commonly used TSF benchmarks, our \Method surpasses or performs on par with SOTA models. For example, on the Solar-Energy dataset, \Method outperforms its closest competitor s-Mamba (0.229 vs. 0.240 in terms of mean-squared error) \textbf{with less parameters, less memory footprint and less computational overhead}. 
\end{itemize}



\section{Related Work} \label{sec:related_work} 

\subsection{Time-series Forecasting} 


\textbf{Transformer-based Models}. 
Transformers, initially introduced by  \cite{vaswani2017attention}, have revolutionized tasks that involve sequence processing and generation, with their self-attention mechanism proving highly effective in capturing long-range dependencies. This architecture, originally designed for natural language processing, has since been adapted to time-series forecasting tasks \citep{ahmed2023transformers}, primarily due to its ability to model complex temporal relationships. \cite{duong2023temporal} demonstrate the efficacy of Transformers in long-term multi-horizon forecasting, addressing the challenge of vanishing correlations over extended horizons.

Recent studies have aimed to address the limitations of standard Transformers in time-series applications.  \cite{foumani2024improving} propose enhanced positional encodings to improve the positional awareness of the Transformer architecture in multivariate time-series classification. \cite{lim2021temporal} propose a Transformer architecture to make use of a complex mix of inputs. \cite{wang2024graphformer} present Graphformer, a model that replaces traditional convolutional layers with dilated convolutional layers, thereby improving the efficiency of capturing temporal dependencies across multiple variates in a graph-based framework.

Despite their unprecedented successes in natural language processing tasks, Transformer models  face several challenges when applied to other time-series domains. One key limitation is their content-based attention mechanism, which struggles to detect crucial temporal dependencies, particularly in cases where dependencies weaken over time or when strong seasonal patterns are present \citep{woo2022etsformer}. Additionally, Transformers suffer from the quadratic complexity of the attention mechanism, which increases computational costs and memory usage, for long input sequences \citep{wen2022transformers}. 

To address these issues, several studies have proposed modifications to the self-attention mechanism. For instance, \cite{zhou2021informer} introduce Informer that employs a sparsified self-attention operation to lower computational complexity and improve long-term forecasting efficiency. Similarly,  \cite{wu2021autoformer} propose Autoformer, which relies on an auto-correlation-based self-attention to better capture temporal dependencies.


\textbf{Linear Models}. 
Linear models are another popular approach in TSF due to their simplicity and efficiency \citep{benidis2022deep}. 
\cite{oreshkin2019n} propose a stacked MLP based architecture with residual links. 
\cite{challu2023nhits} improve this architecture with multi-rate data sampling and hierarchical interpolation for effectively modeling extra long sequences. 
\cite{zeng2023transformers} analyze Transformers for TSF and found that simple linear mappings can outperform Transformer models especially when the data has strong periodic patterns. 
\cite{chen2023tsmixer} introduce another notable linear approach, TSMixer, which leverages an all-MLP architecture to efficiently incorporate cross-variate and auxiliary information. 
\cite{zhang2022less} propose LightTS which is tailored towards efficiently handling very long input series in multivariate TSF. 
\cite{wang2024lightweight} propose time-series Multi-layer Perceptron (MLP), which improves forecasting performance by incorporating domain-specific knowledge into the MLP architecture. 

While linear models with MLPs are simpler architectures and faster compared to Transformer-based models, they face several limitations. These models generally struggle with non-linear dependencies and tend to underperform in scenarios involving highly volatile or non-stationary patterns \citep{chen2023tsmixer}. Moreover, compared to Transformer-based models, linear architectures are less effective at capturing global dependencies. 
This limitation necessitates longer input sequences to achieve comparable forecasting performance, which can increase the computational cost \citep{yi2024frequency}.

\new{
\textbf{Multi-scale Models}. The literature has witnessed many multi-scale time-series models using different architectures. For instance, exploiting the Transformer architecture, Pyraformer \citep{liu2022pyraformer} constructs a multi-resolution pyramidal graph to capture long-range dependencies, while Crossformer \citep{zhang2023crossformer} utilizes Dimension-Segment-Wise embeddings to explicitly model information at varying granularities. Beyond pure attention, TimesNet \citep{wu2023timesnet} transforms 1D series into 2D to capture intra- and inter-period variations, and convolutional approaches like MICN \citep{wang2023micn} and SCINet \citep{liu2022scinet} leverage multi-branch  convolutions or recursive downsampling to fuse local and global contexts. While these methods achieve multi-scale processing through explicit structural hierarchies or downsampling operations, our approach extends this capability to Mamba by leveraging the inherent discretization properties of the sampling rate $\Delta$, enabling multi-resolution feature extraction without architectural downsampling.
}

\subsection{Mamba Models} 
Mamba \cite{gu2023mamba} is a recent sequence model based on State Space Models (SSMs) \cite{gu2021efficiently,smith2022simplified}. Due to its promise of better efficiency-performance trade-off, Mamba has quickly attracted interest from researchers across different domains \cite{qu2024survey,xu2024survey}. 
Mamba's ability to perform content-based reasoning in linear complexity to the sequence length with its hardware-aware algorithm, made it an attractive alternative to the Transformer models. Several works have explored its application to time-series forecasting tasks. \cite{wang2024mamba} propose S-Mamba, which relies on a bidirectional Mamba layer to capture inter-variate dependencies and an MLP to extract temporal dependencies. 
Their model achieves SOTA performance while being faster than Transformer-based alternatives.

\begin{figure*}[hbt!]
    \centering
    \includegraphics[width=0.99\textwidth]{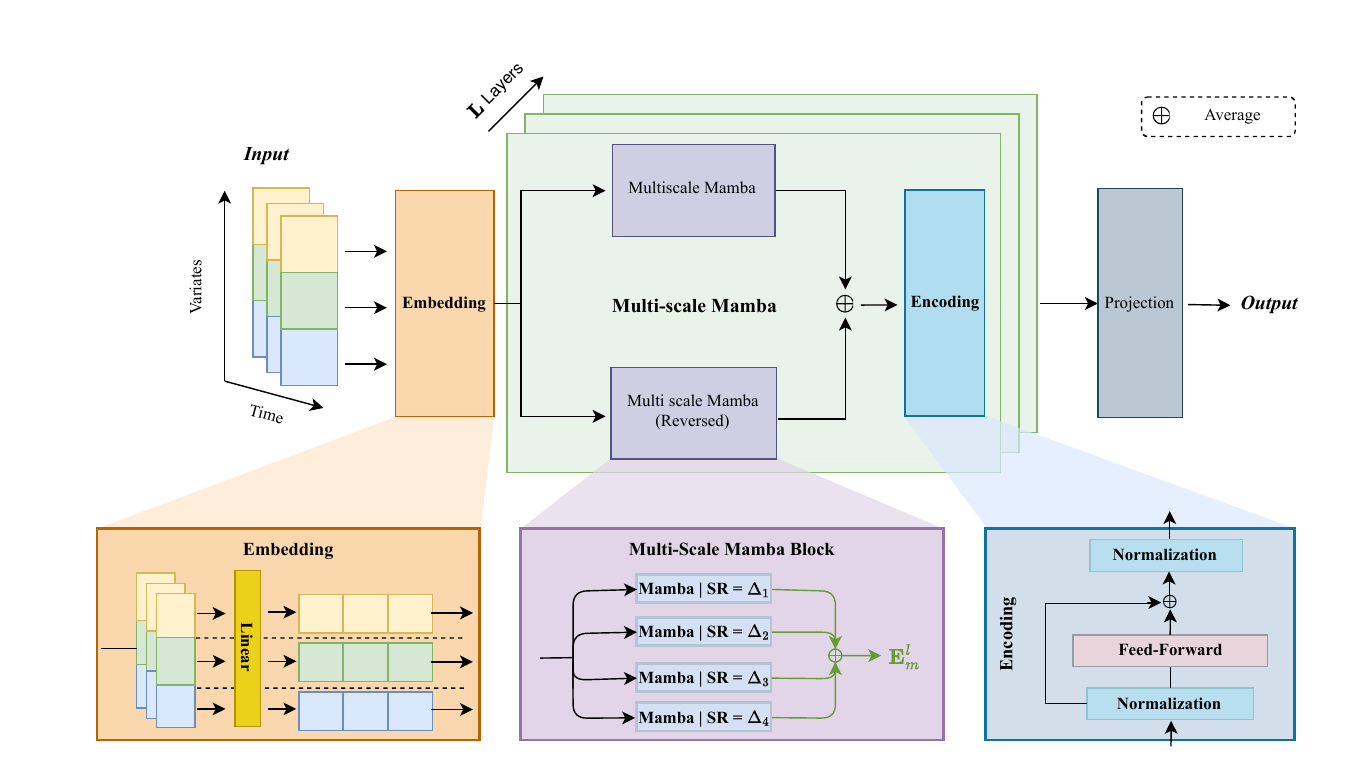}
    \caption{An overview of the proposed method. \Method processes the time-series data at different sampling rates to better capture the multi-scale nature of the input signal. This is achieved by processing and updating the embeddings with different sampling rates (SR). \new{In the input visualization, each square represents a data point for a single variate at a specific time step, with the axes representing the Variates and Time dimensions, respectively.}}   \vspace*{0.2cm}
    \label{fig:overview}
\end{figure*}


\section{Preliminaries and Background}

\subsection{Problem Formulation}

Time-series forecasting is the problem of estimating future $T$ values $\mathbf{Y}_{t+1:t+T}=\{\mathbf{x}_{t+1}, \mathbf{x}_{t+2}, ..., \mathbf{x}_{t+T}\}\in \mathbb{R}^{F\times D}$ of a multi-variate time-series data given its recent $L$ values $\mathbf{X}_{t-L:t} = \{\mathbf{x}_{t-L}, \mathbf{x}_{t-L+1}, ..., \mathbf{x}_t\} \in \mathbb{R}^{L\times D}$ as input. The task is to find the mapping $f$:
\begin{equation}
    f: \mathbf{X}_{t-L:t} \rightarrow \mathbf{Y}_{t+1:t+T},
\end{equation}
which is represented by a deep network whose parameters are estimated from a training dataset.

\subsection{State Space Models (SSMs)}
\label{sect:ssms}

SSMs \cite{gu2021efficiently,smith2022simplified} are sequence models which use a latent state space representation for representing a mapping between a time-series input $x(t)$ and the output $y(t)$ (considering a single-variate setting to simplify notation):
\begin{eqnarray}
    \delta h(t)&=& \textbf{A}h(t)+\textbf{B}x(t),\\
    y(t) &= & \textbf{C}h(t),
\end{eqnarray}
where $\mathbf{A}, \mathbf{B}, \mathbf{C}$ are matrices with learnable values; $h(t)$ is the latent (state) representation; and $\delta h(t)$ is the update for the latent space with the current input, $x(t)$. This continuous formulation is transformed into a discrete model with a sampling rate $\Delta$ as follows:
\begin{eqnarray}
    h_t &= & \hat{\textbf{A}}h_{t-1}+\hat{\textbf{B}}x_t,\\
    y_t &= & {\textbf{C}}h_t,
\end{eqnarray}
where $\hat{\textbf{A}}, \hat{\textbf{B}}, {\textbf{C}}$ are derived by the chosen sampling function (e.g., for zeroth-order hold sampling, $\hat{\textbf{A}} = \exp(\Delta \mathbf{A})$, $\hat{\mathbf{B}} = (\Delta \mathbf{A})^{-1}(\exp(\Delta \mathbf{A}-I)\cdot \Delta \mathbf{B}$ \cite{gu2023mamba}).

SSMs have been recently extended to work more efficiently through the use of a convolutional approach \cite{gu2021combining} and more effectively through better initialization \cite{fu2022hungry}. Moreover, by relating $\hat{\textbf{A}}, \hat{\textbf{B}}, {\textbf{C}}$ to the input, the Mamba model \cite{gu2023mamba} provides comparable or better performance than its Transformer-based counterparts.


\section{Methodology: \Method}

In this section, we describe our \Method in detail. For multi-scale temporal processing, \Method essentially leverages multiple Mamba blocks with different $\Delta$ working in parallel. See Figure \ref{fig:overview} for an overview.

\new{\textbf{Influence of Sampling Rate on Temporal Resolution.} By employing multiple Mamba blocks, each parameterized with a different sampling rate ($\Delta$), our model is designed to intrinsically capture features at multiple, distinct temporal resolutions (or time scales): Overall, large $\Delta$ induces shorter memory and lower temporal resolution whereas small $\Delta$ facilitates long-memory, high-resolution behavior. See Section \ref{sect:theoretical_motivation} for a more in-depth argument.}

\subsection{Embedding Layer}
Following prior work \cite{grazzi2024mamba,liu2023itransformer}, we first transform the input time-series data through an embedding layer (Figure \ref{fig:overview}). Given an input sequence $\mathbf{X} \in \mathbb{R}^{L \times D}$ with $L$ time steps and $D$ variables, we apply a linear transformation along the temporal dimension:
\begin{equation}
\mathbf{E} = \mathrm{Embedding}(\mathbf{X}) \in \mathbb{R}^{D_e \times D},
\end{equation}
where $D_e$ is the embedding dimension. This transformation 
maps each time-series from length $L$ to length $D_e$ while preserving the number of variables $D$, thus enabling us to deal with fixed-length tokens instead of variable input sequence length $L$.  

\subsection{Multi-scale Mamba Layer}
\label{sect:ms_mamba}

As summarized in Section \ref{sect:ssms}, SSMs, Mamba and their variants process time at one learnable sampling rate, $\Delta$. 
Our architecture \Method addresses this gap by processing the input at different sampling rates $\Delta_1$, $\Delta_2$, ..., $\Delta_s$. This is achieved by combining multiple Mamba blocks with different sampling rates as follows:
\begin{equation}
    \mathbf{E}^{l}_m = \mathrm{Avg}(\mathrm{Mamba}(\mathbf{E}^{l}; \Delta_1), ..., \mathrm{Mamba}(\mathbf{E}^{l}; \Delta_n)),
\end{equation}
where $\mathbf{E}^{l}$ is the output of the embedding layer at layer $l$. 

\new{Note that each mamba block contains an internal normalization layer, so our averaging operation fuses the already-normalized outputs from each scale.} \new{Furthermore, keep in mind that the sampling rate $\Delta$ is an internal SSM parameter and not an input downsampling operation. All parallel Mamba blocks receive the same input $\mathbf{E}^{l}$ and produce identically-sized outputs, allowing for direct fusion without any temporal alignment.} 

We explore three different strategies for obtaining $\Delta_i$:
\begin{enumerate}
    \item \textbf{Fixed temporal scales}, where $\Delta_1$ is kept learnable (as in the original Mamba model) but $\Delta_2, \Delta_3, ..., \Delta_n$ are taken as multiples of $\Delta_1$:
    \begin{equation}\label{eq:fixed_scales}
        \Delta_i = \alpha_i\times \Delta_1, \quad\quad i\in \{2, ..., n\},
    \end{equation}
    where $\alpha_i$ are hyper-parameters.

    \item \textbf{Learnable temporal scales}, where all $\Delta_i$ are defined as learnable variables as in the original Mamba model.

    \item \textbf{Dynamic temporal scales}, where all $\Delta_i$ are estimated through a Multi-layer Perceptron:
    \begin{equation}\label{eq:MLP}
        \Delta_i = \mathrm{MLP}(\mathrm{Flatten}(\mathbf{E}^l)),
    \end{equation}
    where $\mathrm{Flatten}(\cdot)$ reshapes the input tensor $\mathbf{E}^l \in \mathbb{R}^{L\times D_e}$ into a vector of dimension $L\cdot D_e$, and $\mathrm{MLP}(\cdot)$ consists of two linear layers with a ReLU activation in between: $\mathrm{MLP}(\mathbf{x}) = \mathbf{W}_2\max(0, \mathbf{W}_1\mathbf{x} + \mathbf{b}_1) + \mathbf{b}_2$, mapping the flattened input (\new{$\mathbf{x}$, representing $\mathrm{Flatten}(\mathbf{E}^l)$ in Equation \ref{eq:MLP}}) to $n$ different sampling rates.
    
\end{enumerate}

To improve the effectiveness of sequential processing, we employ our Multi-scale Mamba module in both directions as illustrated in Figure \ref{fig:overview}, following prior work (e.g., \cite{grazzi2024mamba,zhu2024vision}).

\new{By using different sampling rates $\Delta_i$, each Mamba block generates different discrete-time SSM parameters, allowing each block to specialize in capturing temporal dependencies at a distinct scale. Section \ref{sect:theoretical_motivation} for a more in-depth argument.} 

\subsection{Normalization, Feed-Forward Network and Projection}

%
The output of the Multi-scale Mamba Layer ($\mathbf{E}^l_m$) passes through Layer Normalization, a multi-layer perceptron (MLP -- with two layers with the ReLU nonlinearity) to obtain the embeddings for the next layer ($\mathbf{E}^{l+1} \in \mathbb{R}^{L\times D_e}$):
\begin{equation}
    \mathbf{E}^{l+1} = \mathrm{MLP}(\mathrm{LayerNorm}(\mathbf{E}^l_m)).
\end{equation}
After the last ($N^{th}$) encoder block ($\mathbf{E}^{N} \in \mathbb{R}^{L\times D_e}$), a linear projection layer is applied to map the embedding dimension to prediction length to obtain the final prediction ($\hat{y} \in \mathbb{R}^{F \times D}$): 
\begin{equation}
    \hat{y} = \mathrm{Linear(\mathbf{E}^{N})}.
\end{equation}

\subsection{Training Objective}

The model is trained to minimize the Mean Square Error (MSE) between the predicted values and the ground truth:
\begin{equation}
\mathcal{L} = \frac{1}{F\times D}\sum_{i=1}^F\sum_{j=1}^D(\hat{y}_{i,j} - y_{i,j})^2,
\label{mse}
\end{equation}
where $\hat{y}_{i,j}$ and $y_{i,j}$ are the predicted and ground truth values for the $i$-th time step and $j$-th variable, respectively; $F$ is the forecast horizon; and $D$ is the number of variables. The model parameters are optimized using the Adam optimizer \cite{kingma2014adam} -- see the Suppl. Mat. for more details about the training and experimental details.


\section{Experiments}

Unless otherwise stated, \Method refers to \Method with learnable temporal scales.

\subsection{Experimental Details}

\noindent\textbf{Datasets}.
To evaluate our proposed \Method, we conduct extensive experiments on thirteen real-world time-series forecasting benchmark datasets. The datasets are grouped into three categories for easier comparison. 
\textbf{(1)} Traffic-related datasets, which include Traffic \cite{wu2021autoformer} and PEMS \cite{chen2001freeway}. The Traffic dataset consists of hourly road occupancy rates from the California Department of Transportation, consisting of data collected from 862 sensors on San Francisco Bay area freeways from January 2015 to December 2016. PEMS datasets are complex spatial-temporal datasets for California's public traffic networks, includes four subsets (PEMS03, PEMS04, PEMS07, PEMS08), similar to SCINet. These traffic-related datasets have many periodic features.
\textbf{(2)} ETT (Electricity Transformer Temperature) datasets \cite{zhou2021informer}, which contain load and oil temperature data from electricity transformers, collected between July 2016 and July 2018. This group includes four subsets: ETTm1, ETTm2, ETTh1, and ETTh2, which have fewer variables and show less regularity compared to traffic datasets.
\textbf{(3)} Other datasets: Electricity \cite{wu2021autoformer}, Exchange \cite{wu2021autoformer}, Weather \cite{wu2021autoformer}, and Solar-Energy \cite{lai2018modeling}. The Electricity dataset includes the hourly electricity usage of 321 customers from 2012 to 2014. 
Solar-Energy dataset contains solar power generation data from 137 PV plants in Alabama in 2006, recorded at 10 minute resolution.  The Weather dataset includes 21 meteorological indicators also recorded at 10 minute resolution from the Max Planck Biogeochemistry Institute’s Weather Station in 2020. Exchange dataset compiles daily exchange rates for eight countries from 1990 to 2016. The prior two datasets of this category contain many features most of which are periodic, the last two datasets have fewer primarily aperiodic features. 

\textbf{See the Suppl. Mat for more details on the datasets.}

\begin{table*}[htb!]
  \centering
  \renewcommand{\multirowsetup}{\centering}
    \caption{Experiment 1: Quantitative results on \textbf{traffic-related datasets}. The lookback length $L$ is set to 96 and the forecast length $T$ is set to 12, 24, 48, 96 for PEMS and 96, 192, 336, 720 for Traffic. Top results are highlighted in \first{bold} while the second bests are \second{underlined}. \label{tab:results_traffic}}
  \resizebox{0.98\linewidth}{!}{
  \begin{footnotesize}
  \setlength{\tabcolsep}{2.6pt} 
  \begin{tabular}{c|c|c|c|c|c|c|c|c|c|c|c|c}
    \toprule
    Dataset & $T$ & {\Method} \textbf{(Ours)} & {S-Mamba} & {{iTransformer}} & {RLinear} & {PatchTST} & {{Crossformer}} & {TiDE} & {TimesNet} & {DLinear} & {{FEDformer}} &  {{Autoformer}} \\
    \toprule
    \multirow{5}{*}{\rotatebox{90}{\textbf{Traffic}}} 
    & 96    & \first{0.375} & \second{0.382} & 0.395 & 0.649 & 0.462 & 0.522 & 0.805 & 0.593 & 0.650 & 0.587 & 0.613\\
    & 192   & \first{0.384} & \second{0.396} & 0.417 & 0.601 & 0.466 & 0.530 & 0.756 & 0.617 & 0.598 & 0.604 & 0.616\\
    & 336   & \first{0.408} & \second{0.417} & 0.433 & 0.609 & 0.482 & 0.558 & 0.762 & 0.629 & 0.605 & 0.621 & 0.622 \\
    & 720   & \first{0.442} & \second{0.460} & 0.467 & 0.647 & 0.514 & 0.589 & 0.719 & 0.640 & 0.645 & 0.626 & 0.660 \\
    \cdashline{2-13}[.4pt/1pt]
    & Avg  & \first{0.402} & \second{0.414} & 0.428 & 0.626 & 0.481 & 0.550 & 0.760 & 0.620 & 0.625 & 0.610 & 0.628\\
    \midrule

    \multirow{5}{*}{\rotatebox{90}{\textbf{PEMS03}}} 
    & 12   & \second{0.066} & \first{0.065} & 0.071 & 0.126 & 0.099 & 0.090 & 0.178 & 0.085 & 0.122 & 0.126 & 0.272 \\
    & 24   & \first{0.087} & \first{0.087} & \second{0.093} & 0.246 & 0.142 & 0.121 & 0.257 & 0.118 & 0.201 & 0.149 & 0.334\\
    & 48   & \second{0.133} & \second{0.133} & \first{0.125} & 0.551 & 0.211 & 0.202 & 0.379 & 0.155 & 0.333 & 0.227 & 1.032\\ 
    & 96   & \second{0.201} & \second{0.201} & \first{0.164} & 1.057 & 0.269 & 0.262 & 0.490 & 0.228 & 0.457 & 0.348 & 1.031\\
    \cdashline{2-13}[.4pt/1pt]
    & Avg   & \second{0.122} & \second{0.122} & \first{0.113} & 0.495 & 0.180 & 0.169 & 0.326 & 0.147 & 0.278 & 0.213 & 0.667\\
    \midrule
    
    \multirow{5}{*}{\rotatebox{90}{\textbf{PEMS04}}} 
    & 12   & \first{0.072} & \second{0.076} & 0.078 & 0.138 & 0.105 & 0.098 & 0.219 & 0.087 & 0.148 & 0.138 & 0.424 \\
    & 24   & \first{0.083} & \second{0.084} & 0.095 & 0.258 & 0.153 & 0.131 & 0.292 & 0.10 & 0.224 & 0.177 & 0.459 \\
    & 48   & \first{0.099} & \second{0.115} & 0.120 & 0.572 & 0.229 & 0.205 & 0.409 & 0.136 & 0.355 & 0.270 & 0.646 \\
    & 96   & \first{0.121} & \second{0.137} & 0.150 & 1.137 & 0.291 & 0.402 & 0.492 & 0.190 & 0.452 & 0.341 & 0.912 \\
    \cdashline{2-13}[.4pt/1pt]
    & Avg   & \first{0.094} & \second{0.103} & 0.111 & 0.526 & 0.195 & 0.209 & 0.353 & 0.129 & 0.295 & 0.231 & 0.610 \\
    \midrule
    
    \multirow{5}{*}{\rotatebox{90}{\textbf{PEMS07}}} 
    & 12   & \first{0.060} & \second{0.063} & 0.067 & 0.118 & 0.095 & 0.094 & 0.173 & 0.082 & 0.115 & 0.109 & 0.199 \\
    & 24   & \first{0.075} & \second{0.081} & 0.088 & 0.242 & 0.150 & 0.139 & 0.271 & 0.101 & 0.210 & 0.125 & 0.323 \\
    & 48   & \first{0.091} & \second{0.093} & 0.110 & 0.562 & 0.253 & 0.311 & 0.446 & 0.134 & 0.398 & 0.165 & 0.390 \\
    & 96   & \first{0.109} & \second{0.117} & 0.139 & 1.096 & 0.346 & 0.396 & 0.628 & 0.181 & 0.594 & 0.262 & 0.554 \\
    \cdashline{2-13}[.4pt/1pt]
    & Avg   & \first{0.084} & \second{0.089} & 0.101 & 0.504 & 0.211 & 0.235 & 0.380 & 0.124 & 0.329 & 0.165 & 0.367 \\
    \midrule
    
    \multirow{5}{*}{\rotatebox{90}{\textbf{PEMS08}}}
    & 12   & \first{0.073} & \second{0.076} & 0.079 & 0.133 & 0.168 & 0.165 & 0.227 & 0.112 & 0.154 & 0.173 & 0.436 \\
    & 24   & \first{0.098} & \second{0.104} & 0.115 & 0.249 & 0.224 & 0.215 & 0.318 & 0.141 & 0.248 & 0.210 & 0.467 \\
    & 48  & \first{0.154} & \second{0.167} & 0.186 & 0.569 & 0.321 & 0.315 & 0.497 & 0.198 & 0.440 & 0.320 & 0.966 \\
    & 96   & \second{0.236} & 0.245 & \first{0.221} & 1.166 & 0.408 & 0.377 & 0.721 & 0.320 & 0.674 & 0.442 & 1.385 \\
    \cdashline{2-13}[.4pt/1pt]
    & Avg   & \first{0.140} & \second{0.148} & 0.150 & 0.529 & 0.280 & 0.268 & 0.441 & 0.193 & 0.379 & 0.286 & 0.814 \\
    \bottomrule
  \end{tabular}
  \end{footnotesize}
  }
\end{table*}

\noindent\textbf{Compared Methods}.
We compare our model with 10 state-of-the-art (SOTA) time-series forecasting models belonging to 4 different model families: (1) Mamba based models: S-Mamba \cite{wang2024mamba}; (2) Transformer based models: iTransformer \cite{liu2023itransformer}, PatchTST \cite{nie2022time}, Crossformer \cite{zhang2023crossformer}, FEDformer \cite{zhou2022fedformer}, Autoformer \cite{wu2021autoformer}; (3) Linear based models: TiDE \cite{das2023long}, DLinear \cite{zeng2023transformers}, RLinear \cite{li2023revisiting}; and (4) Temporal Convolution based models: TimesNet \cite{wu2023timesnet}.  The following provides a brief overview of these models:
\begin{itemize}
    \item S-Mamba \cite{wang2024mamba} employs a bidirectional Mamba encoder block to capture inter-variate correlations and a feed forward network temporal dependency encoding layer to learn temporal sequence dynamics. This novel approach is the current SOTA model for TSF task and forms the foundation of our proposed method. 
    \item iTransformer \cite{liu2023itransformer} inverts the order of sequence processing by first analyzing each individual variate separately and then merging the information across all variates.
    \item PatchTST \cite{nie2022time} divides the time-series into sub-series patches treated as input tokens. It leverages channel-independent shared embeddings and weights for efficient representation learning. 
    \item Crossformer \cite{zhang2023crossformer} embeds multivariate time-series into a 2D vector array preserving time and dimension information and introduces two-stage attention to capture both cross-time and cross-dimension dependencies.
    \item FEDformer \cite{zhou2022fedformer} is a frequency-enhanced Transformer that uses trend-seasonality decomposition and exploit sparse representations, Fourier transform, of time-series data to achieve linear complexity to sequence length.
    \item Autoformer \cite{wu2021autoformer} constructs a decomposition architecture that employs traditional sequence decomposition in its inner blocks and utilizes an auto-correlation mechanism.
    \item DLinear \cite{zeng2023transformers} maps trend and seasonality components into predictions via a single linear layer.  
    \item TiDE \cite{das2023long} is a MLP based encoder-decoder model that is best suitable for linear dynamical systems.
    \item RLinear \cite{li2023revisiting} is the current SOTA linear model that introduces reversible normalization and channel independence within a purely linear structure.
    \item TimesNet \cite{wu2023timesnet} proposes a task-general backbone, TimesBlock, that transforms 1D time-series into 2D tensors and uses 2D convolution kernels to capture intra-period and inter-period variations.
\end{itemize}

\noindent\textbf{Training and Implementation Details}.
See the Suppl Mat for training and implementation details, especially the hyperparameters.

\noindent\textbf{Performance Measure}. 
Following the common practice (e.g., \cite{wang2024mamba,liu2023itransformer,nie2022time}), models' performances are compared using the Mean Square Error (MSE) as defined in Equation~\eqref{mse}.

\begin{table*}[htb!]   \vspace*{0.2cm}

  \centering
  \renewcommand{\multirowsetup}{\centering}
    \caption{Experiment 1: Quantitative results on \textbf{ETT Datasets}.
  The lookback length $L$ is set to 96 and the forecast length $T$ is set to 96, 192, 336, 720.
  Top results are highlighted in \first{bold} while the second bests are \second{underlined}. 
  \label{tab:results_ett}}
  \resizebox{0.98\linewidth}{!}{
  \begin{footnotesize}
  \setlength{\tabcolsep}{2.6pt}
  \begin{tabular}{c|c|c|c|c|c|c|c|c|c|c|c|c}
    \toprule
    Dataset & $T$ & {\Method} \textbf{(Ours)} & {S-Mamba} & {{iTransformer}} & {RLinear} & {PatchTST} & {{Crossformer}} & {TiDE} & {TimesNet} & {DLinear} & {{FEDformer}} &  {{Autoformer}} \\ 
    \toprule
    \multirow{5}{*}{\rotatebox{90}{\textbf{ETTm1}}} 
    & 96  & \first{0.326} & 0.333 & 0.334 & 0.355 & \second{0.329} & 0.404 & 0.364 & 0.338 & 0.345 & 0.379 & 0.505 \\
    & 192 & \second{0.371} & 0.376 & 0.377 & 0.391 & \first{0.367} & 0.450 & 0.398 & 0.374 & 0.380 & 0.426 & 0.553 \\
    & 336 & \second{0.406} & 0.408 & 0.426 & 0.424 & \first{0.399} & 0.532 & 0.428 & 0.410 & 0.413 & 0.445 & 0.621 \\
    & 720 & \second{0.470} & 0.475 & 0.491 & 0.487 & \first{0.454} & 0.666 & 0.487 & 0.478 & 0.474 & 0.543 & 0.671 \\ 
    \cdashline{2-13}[.4pt/1pt]
    & Avg & \second{0.394} & 0.398 & 0.407 & 0.414 & \first{0.387} & 0.513 & 0.419 & 0.400 & 0.403 & 0.448 & 0.588 \\
    \midrule    

    \multirow{5}{*}{\rotatebox{90}{\textbf{ETTm2}}} 
    & 96  & \first{0.175} & \second{0.179} & 0.180 & 0.182 & \first{0.175} & 0.287 & 0.207 & 0.187 & 0.193 & 0.203 & 0.255 \\ 
    & 192 & \second{0.244} & 0.250 & 0.250 & 0.246 & \first{0.241} & 0.414 & 0.290 & 0.249 & 0.284 & 0.269 & 0.281\\ 
    & 336 & \second{0.306} & 0.312 & 0.311 & 0.307 & \first{0.305} & 0.597 & 0.377 & 0.321 & 0.369 & 0.325 & 0.339 \\ 
    & 720 & \second{0.407} & 0.411 & 0.412 & \second{0.407} & \first{0.402} & 1.730 & 0.558 & 0.408 & 0.554 & 0.421 & 0.433 \\ 
    \cdashline{2-13}[.4pt/1pt] 
    & Avg & \second{0.283} & 0.288 & 0.288 & 0.286 & \first{0.281} & 0.757 & 0.358 & 0.291 & 0.350 & 0.305 & 0.327 \\ 
    \midrule

    \multirow{5}{*}{\rotatebox{90}{\textbf{ETTh1}}} 
    & 96  & \second{0.384} & 0.386 & 0.386 & 0.386 & 0.414 & 0.423 & 0.479 & \second{0.384} & 0.386 & \first{0.376} & 0.449 \\ 
    & 192 & 0.438 & 0.443 & 0.441 & 0.437 & 0.460 & 0.471 & 0.525 & \second{0.435} & 0.436 & \first{0.420} & 0.500 \\
    & 336 & 0.482 & 0.489 & 0.487 & \second{0.479} & 0.501 & 0.570 & 0.565 & 0.491 & 0.481 & \first{0.459} & 0.521 \\
    & 720 & \second{0.493} & 0.502 & 0.503 & \first{0.481} & 0.500 & 0.653 & 0.594 & 0.521 & 0.519 & 0.506 & 0.514 \\
    \cdashline{2-13}[.4pt/1pt]
    & Avg & 0.449 & 0.455 & 0.454 & \second{0.446} & 0.469 & 0.529 & 0.541 & 0.458 & 0.456 & \first{0.440} & 0.496 \\
    \midrule

    \multirow{5}{*}{\rotatebox{90}{\textbf{ETTh2}}} 
    & 96  & \second{0.291} & 0.296 & 0.297 & \first{0.288} & 0.302 & 0.745 & 0.400 & 0.340 & 0.333 & 0.358 & 0.346 \\
    & 192 & \first{0.369} & 0.376 & 0.380 & \second{0.374} & 0.388 & 0.877 & 0.528 & 0.402 & 0.477 & 0.429 & 0.456 \\
    & 336 & \first{0.412} & 0.424 & 0.428 & \second{0.415} & 0.426 & 1.043 & 0.643 & 0.452 & 0.594 & 0.496 & 0.482 \\
    & 720 & \first{0.418} & 0.426 & 0.427 & \second{0.420} & 0.431 & 1.104 & 0.874 & 0.462 & 0.831 & 0.463 & 0.515 \\
    \cdashline{2-13}[.4pt/1pt]
    & Avg & \first{0.373} & 0.381 & 0.383 & \second{0.374} & 0.387 & 0.942 & 0.611 & 0.414 & 0.559 & 0.437 & 0.450 \\    
    \bottomrule
    \end{tabular}
  \end{footnotesize}
  }
\end{table*}

\begin{table*}[htb!]
  \vspace*{0.2cm}

  \centering
  \caption{Experiment 1: Quantitative results on \textbf{Electricity, Exchange, Weather and Solar-Energy Datasets}.
  The lookback length $L$ is set to 96 and the forecast length $T$ is set to 96, 192, 336, 720.
  Top results are highlighted in \first{bold} while the second bests are \second{underlined}. \label{tab:results_other}}
  \renewcommand{\multirowsetup}{\centering}
  \resizebox{0.98\linewidth}{!}{
  \begin{footnotesize}
  \setlength{\tabcolsep}{2.6pt}
  \begin{tabular}{c|c|c|c|c|c|c|c|c|c|c|c|c}
    \toprule
    Dataset & $T$ & {\Method} \textbf{(Ours)} & {S-Mamba} & {{iTransformer}} & {RLinear} & {PatchTST} & {{Crossformer}} & {TiDE} & {TimesNet} & {DLinear} & {{FEDformer}} &  {{Autoformer}} \\
    \toprule
    \multirow{5}{*}{\rotatebox{90}{\textbf{Electricity}}} 
    & 96 & \first{0.138} & \second{0.139} & 0.148 & 0.201 & 0.181 & 0.219 & 0.237 & 0.168 & 0.197 & 0.193 & 0.201 \\
    & 192 & \first{0.157} & \second{0.159} & 0.162 & 0.201 & 0.188 & 0.231 & 0.236 & 0.184 & 0.196 & 0.201 & 0.222 \\
    & 336 & \first{0.174} & \second{0.176} & 0.178 & 0.215 & 0.204 & 0.246 & 0.249 & 0.198 & 0.209 & 0.214 & 0.231 \\ 
    & 720 & \first{0.199} & \second{0.204} & 0.225 & 0.257 & 0.246 & 0.280 & 0.284 & 0.220 & 0.245 & 0.246 & 0.254 \\
    \cdashline{2-13}[.4pt/1pt]
    & Avg & \first{0.167} & \second{0.170} & 0.178 & 0.219 & 0.205 & 0.244 & 0.251 & 0.192 & 0.212 & 0.214 & 0.227 \\
    \midrule

    \multirow{5}{*}{\rotatebox{90}{\textbf{Exchange}}} 
    & 96  & \first{0.086} & \first{0.086} & \first{0.086} & 0.093 & \second{0.088} & 0.256 & 0.094 & 0.107 & \second{0.088} & 0.148 & 0.197 \\
    & 192 & 0.178 & 0.182 & \second{0.177} & 0.184 & \first{0.176} & 0.470 & 0.184 & 0.226 & \first{0.176} & 0.271 & 0.300 \\
    & 336 & 0.326 & 0.332 & 0.331 & 0.351 & \first{0.301} & 1.268 & 0.349 & 0.367 & \second{0.313} & 0.460 & 0.509 \\
    & 720 & \second{0.843} & 0.867 & 0.847 & 0.886 & 0.901 & 1.767 & 0.852 & 0.964 & \first{0.839} & 1.195 & 1.447 \\
    \cdashline{2-13}[.4pt/1pt] 
    & Avg & \second{0.358} & 0.367 & 0.360 & 0.378 & 0.367 & 0.940 & 0.370 & 0.416 & \first{0.354} & 0.519 & 0.613 \\
    \midrule
    
    \multirow{5}{*}{\rotatebox{90}{\textbf{Weather}}} 
    & 96  & \second{0.163} & 0.165 & 0.174 & 0.192 & 0.177 & \first{0.158} & 0.202 & 0.172 & 0.196 & 0.217 & 0.266 \\    
    & 192 & \second{0.213} & 0.214 & 0.221 & 0.240 & 0.225 & \first{0.206} & 0.242 & 0.219 & 0.237 & 0.276 & 0.307 \\
    & 336 & \first{0.270} & 0.274 & 0.278 & 0.292 & 0.278 & \second{0.272} & 0.287 & 0.280 & 0.283 & 0.339 & 0.359 \\
    & 720 & \second{0.349} & 0.350 & 0.358 & 0.364 & 0.354 & 0.398 & 0.351 & 0.365 & \first{0.345} & 0.403 & 0.419 \\
    \cdashline{2-13}[.4pt/1pt]
    & Avg & \first{0.249} & \second{0.251} & 0.258 & 0.272 & 0.259 & 0.259 & 0.271 & 0.259 & 0.265 & 0.309 & 0.338 \\
    \midrule

    \multirow{5}{*}{\rotatebox{90}{\textbf{Sol. Ener.}}} 
    & 96  & \first{0.195} & 0.205 & \second{0.203} & 0.322 & 0.234 & 0.310 & 0.312 & 0.250 & 0.290 & 0.242 & 0.884 \\
    & 192 & \first{0.230} & 0.237 & \second{0.233} & 0.359 & 0.267 & 0.734 & 0.339 & 0.296 & 0.320 & 0.285 & 0.834 \\
    & 336 & \first{0.247} & 0.258 & \second{0.248} & 0.397 & 0.290 & 0.750 & 0.368 & 0.319 & 0.353 & 0.282 & 0.941 \\
    & 720 & \first{0.249} & \second{0.260} & \first{0.249} & 0.397 & 0.289 & 0.769 & 0.370 & 0.338 & 0.356 & 0.357 & 0.882 \\
    \cdashline{2-13}[.4pt/1pt]
    & Avg & \first{0.229} & 0.240 & \second{0.233} & 0.369 & 0.270 & 0.641 & 0.347 & 0.301 & 0.330 & 0.291 & 0.885 \\    
    \bottomrule
  \end{tabular}
  \end{footnotesize}
  }
\end{table*}

\subsection{Experiment 1: Comparison with State-of-the-Art}

Unless stated otherwise, \Method refers to \Method with learnable temporal scales.

\noindent\textbf{Traffic-related Datasets (Table \ref{tab:results_traffic})}. The results in Table \ref{tab:results_traffic} show that \Method provides the best or second best results over all traffic datasets across all forecast lengths. Compared to our baseline model of S-Mamba, \Method delivers significant improvements, especially on the Traffic dataset (0.402 vs. 0.414).

\noindent\textbf{ETT Datasets (Table \ref{tab:results_ett})}. On ETT datasets, as in Table \ref{tab:results_ett}, \Method is typically the second best and the best performing method in ETTh2 and in some configurations of ETTm1 and ETTm2. Compared to S-Mamba, \Method provides better performance. 

\noindent\textbf{Other Datasets (Table \ref{tab:results_other})}. The results in other datasets (Table \ref{tab:results_other}) are in agreement with the results on Traffic-related and ETT datasets: Our \Method performs better than or is generally on par with SOTA methods, and consistently provides better performance than the S-Mamba baseline, especially on the Solar Energy dataset (0.229 vs. 0.240).

\noindent\textbf{Summary:} Our \Method exhibits strong performance across 13 benchmark datasets and diverse problems compared to the baseline method of s-Mamba and the SOTA methods. \new{We observe that the performance gains are particularly pronounced on datasets with distinct, hierarchical temporal patterns (e.g., Traffic and Solar Energy). This supports our hypothesis that assigning distinct Mamba blocks to different sampling rates allows the model to better disentangle and capture these overlapping temporal dynamics compared to single-scale baseline.}

\begin{table*}[hbt!]
    \centering
   \vspace*{0.2cm}
   \caption{Experiment 2: Ablation study on Traffic and Solar Energy datasets. The lookback length $L=96$, while the forecast length $T \in \left\{96, 192, 336, 720\right\}$. $\mathbf{\alpha} = (\alpha_1, \alpha_2, \alpha_3, \alpha_4)$ indicates that the $\Delta_1$ (learnable sampling rate of the base Mamba) is multiplied by these coefficients to obtain the sampling rates for the Mamba blocks (Eq. \ref{eq:fixed_scales}). Top results are highlighted in \first{bold} while the second bests are \second{underlined}. \label{tab:full_ab}}
  \setlength{\tabcolsep}{3.1pt}
    \begin{small}

    \begin{tabular}{l|cccc|cccc}
    \toprule
    {\hfill \hspace*{1cm} Dataset $\Rightarrow$} & \multicolumn{4}{c|}{Traffic} & \multicolumn{4}{c}{Solar Energy}\\
    \cmidrule(lr){1-1}\cmidrule(lr){2-5}\cmidrule(lr){6-9}
    Prediction Length $\Rightarrow$ & 96 & 192 & 336 & 720 & 96 & 192 & 336 & 720\\
    \midrule
    \rowcolor{gray!30!}\multicolumn{9}{c}{\Method with fixed temporal scales}\\
    \midrule
    {\textbf{$\mathbf{\alpha} = (1, 2, 4, 8)$}} 
    & .376 & .392 & \first{.405} & \second{.452} & 
    .196 & \first{.230} & .250 & \first{.249}\\
    
    {$\mathbf{\alpha} = (0.5,1, 1.5, 2)$} 
    & \first{.374} & \second{.389} & .414 & .458 & 
    .197 & .232 & \second{.248} & \second{.251}\\
    
    {$\mathbf{\alpha} = (1, 2, 3, 4)$} 
     & .390 & .403 & .415 & .455 &
     .196 & .232 & .250 & \second{.251}\\
    
    {$\mathbf{\alpha} = (1, 4, 8, 16)$} 
    & .380 & .411 & .421 & .453 &
    .197 & .232 & .250 & \second{.251}\\
    \midrule
    \rowcolor{gray!30!}\multicolumn{9}{c}{\Method with learnable scales}\\
    \midrule
    &  \second{.375} & \first{.384} & \second{.408} & \first{.442} &
    \second{.195} & \first{.230} & \first{.247} & \first{.249}\\
    \midrule
    \rowcolor{gray!30!}\multicolumn{9}{c}{\Method with dynamic scales}\\
    \midrule
    & .376 & .390 & .414 & .454 &
    \first{.194} & \second{.231} & .249 & \second{.251}\\

    \bottomrule
    \end{tabular}
    \end{small}
\end{table*}

\begin{table}[hbt!]
    \centering
    \setlength{\tabcolsep}{.1cm}
    \footnotesize
    \caption{Experiment 2: Effect of the number of scales using the Solar Energy dataset with $T = 96$.}
    \label{tab:effect_of_number_of_scales}
    \resizebox{0.9\linewidth}{!}{
    \begin{tabular}{l|c|c|c|c|c}
        \toprule
        Method / Scale Count & 2 & 3 & 4 & 5 & 6 \\
        \midrule
        \Method & 0.202 & 0.199 & \textbf{{0.196}} & 0.199 & 0.203 \\ \midrule
        
    \end{tabular}
    }
\end{table}

\subsection{Experiment 2: Ablation Analysis}


\textbf{Evaluation of Different Strategies for Setting $\Delta_i$.}
First, we evaluate the performances of the strategies described in Section \ref{sect:ms_mamba}: (i) \Method with fixed temporal scales. (ii) \Method with learnable temporal scales. (iii) \Method with dynamic temporal scales. For this analysis, we tune the hyperparameters in all settings. To keep the number of experiments manageable, we consider one dataset from each category.

The results in Table \ref{tab:full_ab} suggest that using learned temporal scales performs best among the different strategies considered (5 best and 3 second-best results out of 8 settings). Fixed temporal scales provide on par results in many different settings. These results suggest that a more refined search for fixed temporal scales might provide better results. However, this requires more experiments for hyperparameter tuning, which is alleviated by the learnable temporal scales approach.

The dynamic scales strategy generally provides the worst performance, except for the Solar Energy dataset. The subpar performance of the dynamic approach may be owing to the extra learnable parameters introduced by the approach.

\noindent\textbf{Effect of the number of scales.} In this analysis, for the best performing \Method version (\Method with temporal scales), we evaluate the impact of the number of scales considered. As listed in Table \ref{tab:effect_of_number_of_scales}, we observe that having 4 scales provides the best performance overall. 

\begin{figure}[hbt!]
    \centering
    \includegraphics[width=1.0\linewidth]{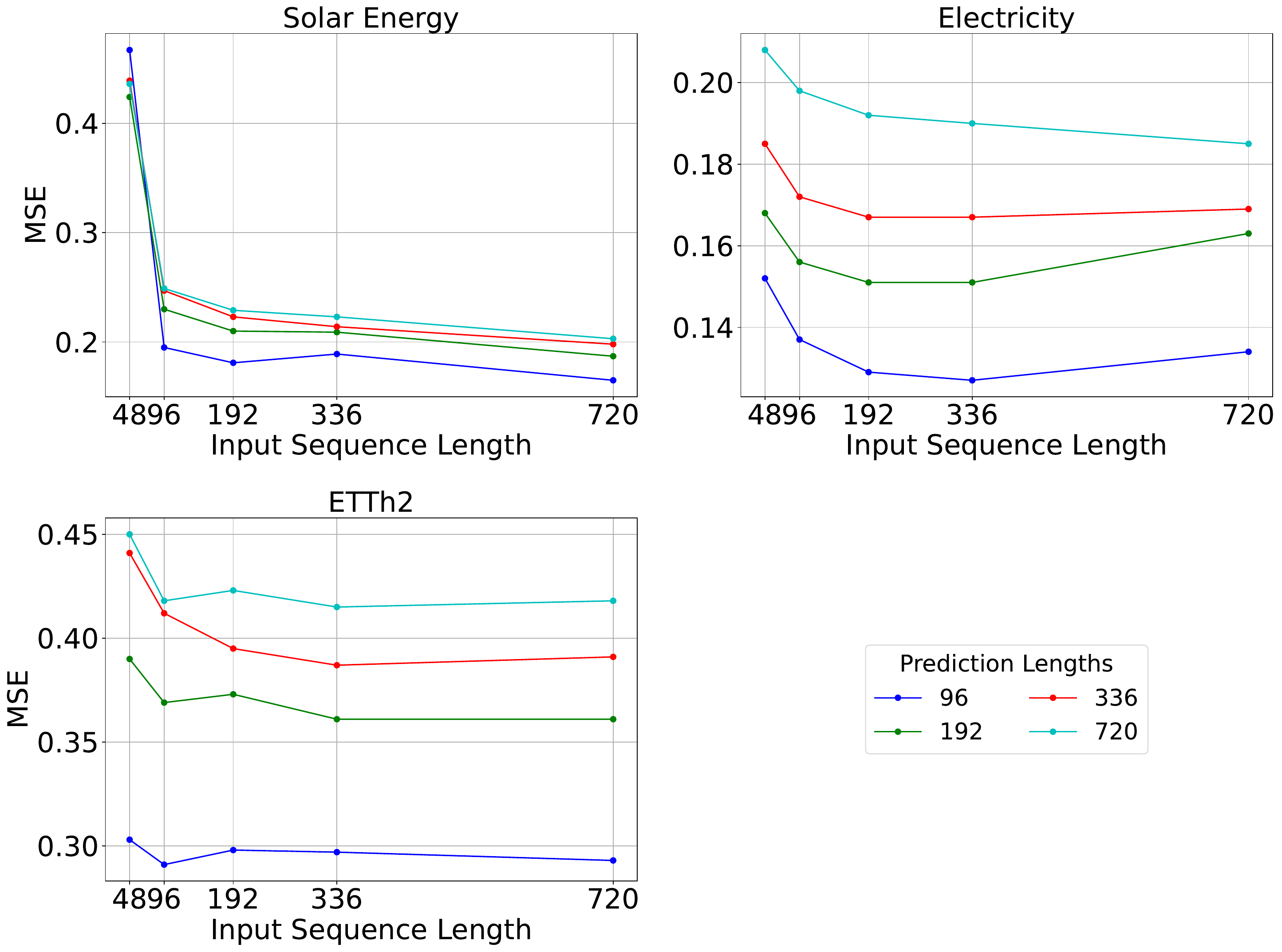}
    \caption{\new{Effect of input length ($L$) on the performance of \Method.}}
    \label{fig:input_length}
\end{figure}

\noindent\new{\textbf{Effect of input length.} In Figure \ref{fig:input_length}, we analyze the impact of the input length on the performance of \Method. We see that larger input length provides better results for the Solar and Electricity datasets whereas an input length of 96 provides the best results for the ETTh2 dataset. Although there is no single input length that works best for all datasets, for fair comparison with the literature, we have used 96 as the input length in all experiments.
}

\textbf{Summary:} Learnable temporal scales provides the best results whilst not requiring any hyperparameter tuning as in the fixed temporal scales approach. Therefore, unless stated otherwise, \Method refers to \Method with learnable temporal scales.

\begin{figure*}[hbt!]
    \centering
    \includegraphics[width=0.99\textwidth]{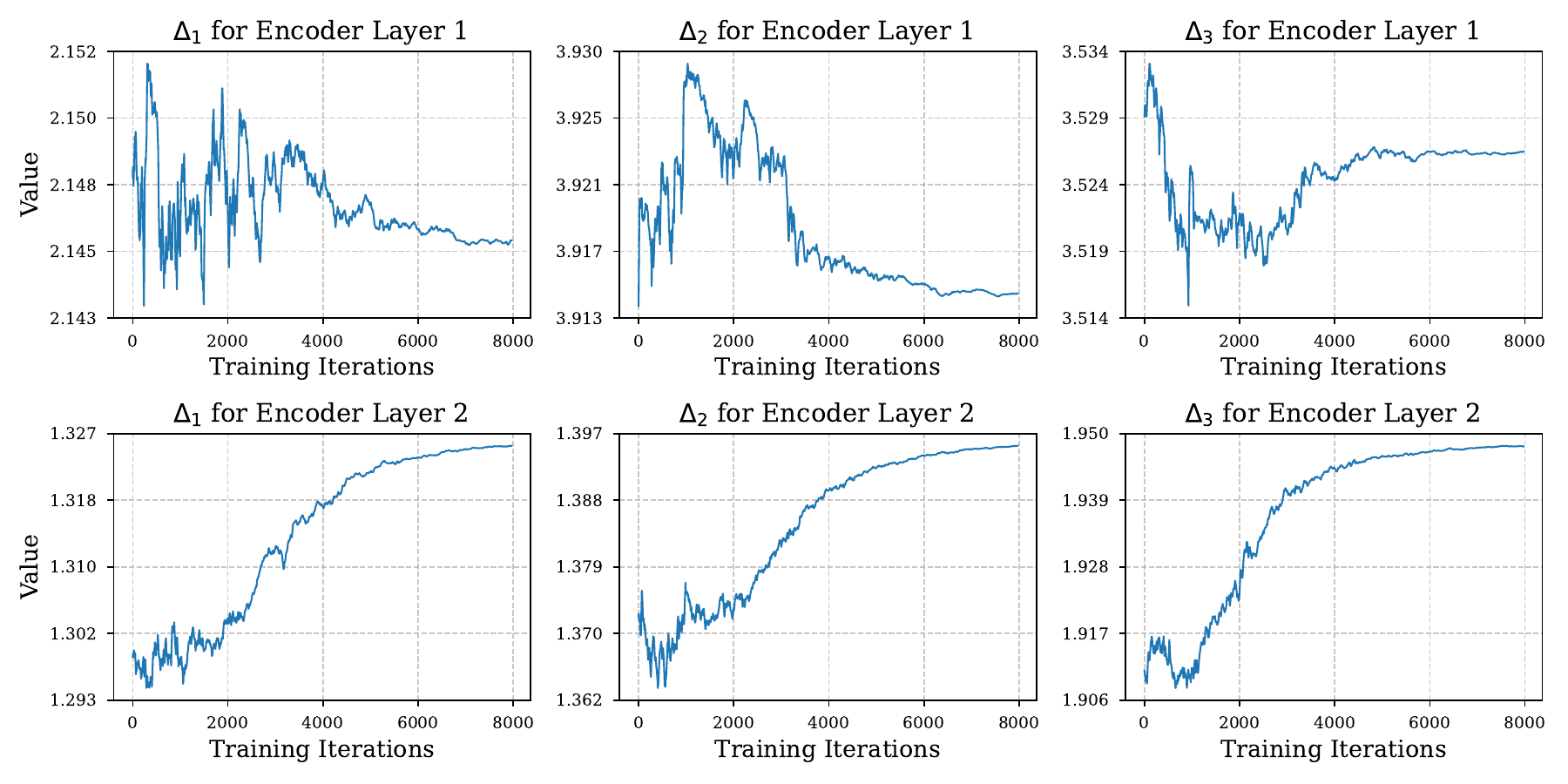}
    \caption{Experiment 3: Values of learnable $\Delta_i$ parameters over time for the Solar Energy dataset where $T=96$ ($\Delta_i$ parameters are initialized randomly between 1 and 4).}
    \label{fig:multiplierPlot}
\end{figure*}

\begin{figure*}[hbt!]
    \centering
    \subcaptionbox{ms-Mamba}{
    \includegraphics[width=0.99\textwidth]{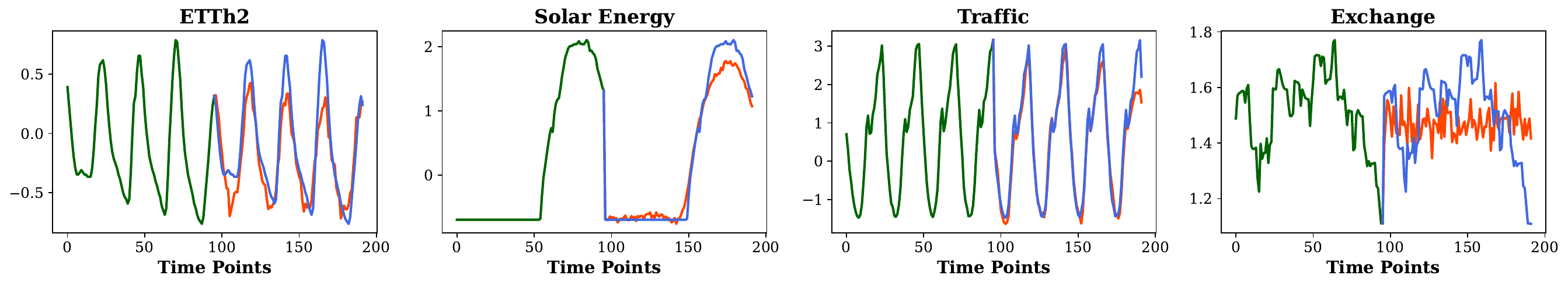}
    }\\
    \vspace{10pt}
    \subcaptionbox{S-Mamba}
    {
    \includegraphics[width=0.99\textwidth]{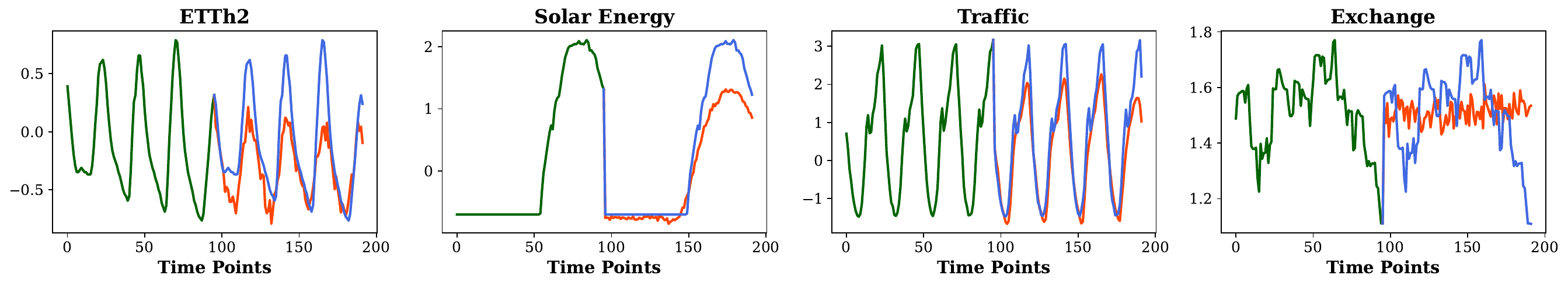}
    }\\
    \vspace{10pt}
    \caption{Experiment 3: Forecast comparison between \Method and S-Mamba on four datasets when the input length is 96 ($L = 96$) and the
forecast length is 96 ($T = 96$). The green line represents the input, the blue line represents the ground truth and the red line represents the forecast.}
    \label{fig:comparisonPlot}
\end{figure*}

\begin{table}[htb!]\vspace*{0.2cm}
 \caption{Experiment 4: Efficiency comparison of \Method and S-Mamba. \del{on one dataset from each category.} The lookback length $L$ is set to 96 and the forecast length $T$ is set to 96, 192, 336, 720. \new{We report results on the extended ETT dataset family to demonstrate intra-category consistency, alongside representative datasets from other categories.}\label{tab:result_perf}}
  \resizebox{1\linewidth}{!}{
  \setlength{\tabcolsep}{2.6pt}
  \begin{tabular}{c|c|cccc|cccc}
    \toprule
    \multicolumn{2}{c|}{Models} & \multicolumn{4}{c|}{{\Method} \textbf{(Ours)}} & \multicolumn{4}{c}{S-Mamba}\\
    \cmidrule(lr){1-2}\cmidrule(lr){3-6}\cmidrule(lr){7-10}  
    \multicolumn{2}{c|}{Metric} & MSE & \#Params & Memory & MACs & MSE & \#Params & Memory & MACs\\
    \toprule
    \multirow{4}{*}{\rotatebox{90}{\textbf{ETTh1}}} 
    & 96  & \textbf{0.384} & \textbf{0.195M} &  \textbf{0.746MB} & \textbf{0.063G} & 0.386 & 1.145M & 4.37MB & 0.400G\\
    & 192  & \textbf{0.438} & \textbf{0.179M} & \textbf{0.685MB} & \textbf{0.057G} & 0.443 & 1.170M & 4.46MB & 0.409G\\
    & 336  & \textbf{0.482} & \textbf{0.270M} & \textbf{1.03MB} & \textbf{0.087G} & 0.489 & 1.207M & 4.60MB & 0.422G\\
    & 720  & \textbf{0.493} & \textbf{0.223M} & \textbf{0.85MB} & \textbf{0.075G} & 0.502 & 1.306M & 4.98MB & 0.457G\\

    \midrule
    \multirow{4}{*}{\rotatebox{90}{\textbf{ETTh2}}} 
    & 96  & \textbf{0.291} & \textbf{0.481M} &  \textbf{1.84MB} & \textbf{0.165G} & 0.296 & 1.150M & 4.40MB & 1.563G\\
    & 192  & \textbf{0.369} & \textbf{0.484M} & \textbf{1.85MB} & \textbf{0.171G} & 0.376 & 1.175M & 4.48MB & 1.580G\\
    & 336  & \textbf{0.412} & \textbf{0.503M} & \textbf{1.92MB} & \textbf{0.180G} & 0.424 & 1.212M & 4.62MB & 1.606G\\
    & 720  & \textbf{0.418} & \textbf{0.552M} & \textbf{2.11MB} & \textbf{0.195G} & 0.426 & 1.311M & 5.00MB & 1.675G\\

    \midrule
    \multirow{4}{*}{\rotatebox{90}{\textbf{ETTm1}}} 
    & 96  & \textbf{0.326} & \textbf{0.160M} & \textbf{0.61MB} & \textbf{0.055G} & 0.333 & 1.15M & 4.37MB & 0.400G\\
    & 192  & \textbf{0.371} & \textbf{0.121M} & \textbf{0.46MB} & \textbf{0.028G} & 0.376 & 1.201M & 17.90MB & 0.109G\\
    & 336  & \textbf{0.406} & \textbf{0.191M} & \textbf{0.73MB} & \textbf{0.065}G & 0.408 & 0.333M & 1.272B & 0.116G\\
    & 720  & \textbf{0.470} & \textbf{0.223M} & \textbf{0.85MB} & \textbf{0.075G} & 0.475 & 0.383M & 1.46MB & 0.133G\\

    \midrule
    \multirow{4}{*}{\rotatebox{90}{\textbf{ETTm2}}} 
    & 96  & \textbf{0.175} & \textbf{0.304M} & \textbf{1.16MB} & \textbf{0.102G} & 0.179 & 1.15M & 4.37MB & 0.400G\\
    & 192  & \textbf{0.244} & \textbf{0.168M} & \textbf{0.64MB} & \textbf{0.058G} & 0.250 & 0.315M & 1.20MB & 0.109G\\
    & 336  & \textbf{0.306} & \textbf{0.132M} & \textbf{0.505MB} & \textbf{0.044}G & 0.312 & 0.333M & 1.272B & 0.116G\\
    & 720  & \textbf{0.407} & \textbf{0.189M} & \textbf{0.72MB} & \textbf{0.063G} & 0.411 & 0.383M & 1.46MB & 0.133G\\
    
    \midrule
    \multirow{4}{*}{\rotatebox{90}{\textbf{Solar En.}}} 
    & 96  & \textbf{0.195} & \textbf{3.958M} & \textbf{15.10MB} & \textbf{16.72G} & 0.205 & 4.643M & 17.71MB & 20.00G\\
    & 192  & \textbf{0.230} & \textbf{2.028M} & \textbf{7.74MB} & \textbf{8.57G} & 0.237 & 4.692M & 17.90MB & 20.21G\\
    & 336  & \textbf{0.247} & \textbf{4.015M} & \textbf{15.31MB} & \textbf{16.99}G & 0.258 & 4.766M & 18.18MB & 20.54G\\
    & 720  & \textbf{0.249} & \textbf{4.113M} & \textbf{15.70MB} & \textbf{17.43G} & 0.260 & 4.963M & 18.93MB & 21.40G\\

    \midrule
    \multirow{4}{*}{\rotatebox{90}{\textbf{Traffic}}} 
    & 96  & \textbf{0.375} & 29.68M & 113.2MB & 403.5G & 0.382 & \textbf{9.186M} & \textbf{35.04MB} & \textbf{125.0G}\\
    & 192  & \textbf{0.384} & 14.94M & 56.99MB & 203.1G & 0.396 & \textbf{9.236M} & \textbf{35.23MB} & \textbf{125.7G}\\
    & 336  & \textbf{0.408} & 29.81M & 113.7MB & 405.2G & 0.417 & \textbf{9.310M} & \textbf{35.51MB} & \textbf{126.7G}\\
    & 720  & \textbf{0.442} & 29.68M & 114.5MB & 407.9G & 0.460 & \textbf{9.507M} & \textbf{36.26MB} & \textbf{129.5G}\\
    \cmidrule(lr){2-10}    
  \end{tabular}
  }
\end{table}

\subsection{Experiment 3: Qualitative Analysis}

\noindent\textbf{Evolution of learned $\alpha$.} In Figure \ref{fig:multiplierPlot}, we investigate how the learned $\Delta_i$ changes throughout training. We observe that the $\Delta_i$ values fluctuate during the initial stages of training. However, they converge to and saturate at certain values.  

\noindent\textbf{Visual results.} Next, we compare the forecasts of our \Method and the baseline s-Mamba method. As shown in Figure \ref{fig:comparisonPlot}, s-Mamba has a tendency to undershoot peak values in ETTh2, Solar Energy and Traffic datasets. \new{This limitation in the s-Mamba model likely stems from its reliance on a single temporal scale, which forces a trade-off between capturing high-frequency variations (peaks) and low-frequency trends. By operating with multiple sampling rates ($\Delta$) simultaneously, \Method avoids this trade-off: branches with different $\Delta$ values can specialize, allowing the model to capture sharp, rapid transitions (via smaller $\Delta$s) without sacrificing the modeling of longer-term trends.} However, both methods struggle on the Exchange dataset, which is known to be a challenging problem for TSF models.

\subsection{Experiment 4: Efficiency Analysis}

In this experiment, we investigate the efficiency of \Method (with learnable temporal scales) in comparison with the baseline s-Mamba. In each dataset, we provide the results of the best performing configurations for each method.

As listed in Table \ref{tab:result_perf}, on the \del{ETTh1}\new{datasets from ETT category} and Solar Energy dataset, we see that \Method provides the best results with less parameters, memory and operations. This is crucial as it shows multiple temporal scales can be utilized with less computational overhead. However, this result is not observed in the Traffic dataset because it contains significantly more variates (862) compared to the ETTh2 (7) and Solar Energy (137) datasets. 

\noindent\textbf{Summary:} \Method provides better performance than the baseline S-Mamba with less parameters, memory and computations on many datasets. However, this is not observed for the \del{ ETTh2 and Solar Energy datasets.} \new{Traffic dataset.}

    


\section{Conclusion}

In this paper, we introduce a novel multi-scale architecture for the problem of time-series forecasting (TSF). Our architecture extends Mamba (or its derivative S-Mamba) where we include several Mamba blocks with different sampling rates to process multiple temporal scales simultaneously. The different sampling rates can either be fixed or learned from the data, which leads to a simple architecture with a multi-scale processing ability. 

Our results on 13 TSF benchmarks over different problem kinds show that our approach provides the best or on par performance compared to the state-of-the-art methods. What is remarkable is that, compared to the baseline model (S-Mamba), \Method provides better results \del{with} \new{and, on many datasets, does so} with  less parameters, memory, and operations. 

\textbf{Limitations and Future Work}. It is a promising research direction to apply \Method for other types of modalities, e.g., text and images. Moreover, \Method can complement other types of deep modules, e.g., scaled-dot-product attention, linear attention. \new{Furthermore, while our  averaging fusion was effective and efficient, exploring more sophisticated, learnable fusion mechanisms (e.g., attention or weighted-averaging) to better exploit the complementarity between scales is a promising direction for future work.}


\section*{Acknowledgment}

This work has been partially funded by the European Union’s Horizon project LEGOFIT - Adaptable technological solutions based on early design actions for the construction and renovation of Energy Positive Homes (Grant agreement ID: 101104058). We also gratefully acknowledge the computational resources kindly provided by METU Image Lab and METU-ROMER, Center for Robotics and Artificial Intelligence.


\appendix

\section{Theoretical Motivation for Multiple $\Delta$s}
\label{sect:theoretical_motivation}

By employing multiple Mamba blocks, each parameterized with a different sampling rate $\Delta_i$, our model is designed to intrinsically capture features at multiple, distinct temporal resolutions (or time scales). Recall from Section \ref{sect:ssms} that:
\begin{align}
    \hat{A} &= \exp(\Delta A), \\
    \hat{B} &= A^{-1} (\exp(\Delta A) - I) \, B \approx (\Delta A)^{-1}(\exp(\Delta A)-I)\,\Delta B,
\end{align}
Thus, the choice of $\Delta$ directly controls the discrete-time dynamics by altering both the transition matrix $\hat{A}$ and the input-response matrix $\hat{B}$. Using multiple SSMs with different $\Delta$ therefore induces multiple characteristic timescales in the latent dynamics.

\subsection{Influence of Small $\Delta$}

When $\Delta$ is small, the discretization stays close to the identity map:
\begin{align}
    \hat{A} = \exp(\Delta A) \approx I + \Delta A.
\end{align}
This implies that the eigenvalues of $\hat{A}$ lie close to~1, leading to slow decay of the hidden state.  
Consequently, the SSM retains information over many discrete time steps, capturing \emph{fine-grained temporal variations} and high-frequency structure.

The input mapping behaves similarly:
\begin{align}
    \hat{B} \approx \Delta B,
\end{align}
so the effective input contribution per time step is small but frequent.  
Taken together, small $\Delta$ yields a high-resolution temporal process: the latent state evolves slowly, and the model is sensitive to rapid changes in the input.

\subsection{Influence of Large $\Delta$}

For large $\Delta$, the discretized transition becomes
\begin{align}
    \hat{A} = \exp(\Delta A),
\end{align}
which strongly contracts the dynamics whenever $A$ has negative real eigenvalues.  
Eigenvalues of $\exp(\Delta A)$ shrink rapidly in magnitude as $\Delta$ increases, causing fast decay of the hidden state.  
This yields a \emph{coarse-grained temporal process}: the model integrates information over broad temporal windows and captures slow trends rather than rapid fluctuations.

The behavior of the input operator reflects this as well.  
Since in the following definition:
\begin{align}
    \hat{B} = A^{-1} (\exp(\Delta A) - I) B,
\end{align}
large $\Delta$ causes $(\exp(\Delta A)-I)$ to amplify modes of $A$ with large negative real parts, producing a stronger input influence per update step.  
This leads the model to respond primarily to aggregated (low-frequency) input structure.

Overall, large $\Delta$ induces shorter memory and lower temporal resolution, complementing the long-memory, high-resolution behavior produced by small $\Delta$.

\section{More Details about the Experimental Setup}
\label{sec:appendix_setup} 

\subsection{Dataset Statistics}

In this section (in Table \ref{tab:datasets}), we summarize the properties of the datasets used in the experiments.

\begin{table}
\caption{13 public benchmark datasets used in the paper.}
    \label{tab:datasets}
    \centering\footnotesize
    \begin{tabular}{c|cccc}
        \hline
         Datasets & Variates & Timesteps & Resolution & \\
         \hline
         Traffic & 862 & 17,544 & 1 hour & \\
         PEMS03 & 358 & 26,209 & 5 minutes & \\
         PEMS04& 307 & 16,992 & 5 minutes & \\
         PEMS07& 883 & 28,224 & 5 minutes & \\
         PEMS08& 170 & 17,856 & 5 minutes & \\
         ETTm1 \& ETTm2 & 7 & 17,420 & 15 minutes & \\
         ETTh1 \& ETTh2 & 7 & 69,680 & 1 hour & \\
         Electricity & 321 & 26,304 & 1 hour & \\
         Exchange & 8 & 7,588 & 1 day & \\
         Weather & 21 & 52,696 & 10 minutes & \\
         Solar Energy & 137 & 52,560 & 10 minutes & \\
         \hline
    \end{tabular}
\end{table}

\subsection{Training and Implementation Details}

For a fair comparison, we follow the experimental settings of S-Mamba \cite{wang2024mamba} where input sequence length is fixed at 96 time steps and prediction lengths for training and testing are fixed at 96, 192, 336 and 720 for each dataset. \new{To ensure a fair comparison, we adopt the specific epoch settings from the S-Mamba baseline: the maximum number of training epochs is set to 5 for PEMS03, Electricity, and Weather, and 10 for the remaining datasets (Traffic, PEMS04, PEMS07, PEMS08, the ETT category, Exchange, and Solar Energy).} We tuned the learning rate, batch size, the number of encoder layers, embedding dimension and FFN dimension in our experiments -- see Table \ref{tab:hyperparams} for the ranges considered. For datasets with fewer variates, we explored smaller embedding and FFN dimensions, whereas for datasets with more variates, we explored higher dimensions. For alignment of results with the original experimental set-up, we did not explicitly tune the random seed. During the {hyperparameter optimization phase}, we measure the mean-squared error (MSE) performance on the validation set and use early stopping. {\tt \Method} models are implemented in {\tt PyTorch} framework. Our experiments are conducted on a cluster containing NVIDIA RTX 3090 and NVIDIA RTX A6000 GPUs. 

\begin{table}[hbt!]
\caption{Model Hyperparameters and Their Ranges}
\label{tab:hyperparams}
\centering\footnotesize
\begin{tabular}{lc}
\toprule
\textbf{Hyperparameter} & \textbf{Range} \\
\midrule
Learning Rate & ($1e^{-5}$, $1e^{-3}$) \\
Batch Size & \{16, 32, 64\}  \\
Encoder Layers & \{1, 2, 4\} \\
Embedding Dimension & \{32, 64, 80, 128, 256, 512, 768, 1024\} \\
FFN Dimension & \{32, 64, 80, 128, 256, 512, 768, 1024\} \\
\bottomrule
\end{tabular}
\end{table}

\subsection{Additional Visualizations of Multi-scale Patterns}
\label{sec:appendix_vis}

\begin{figure}[t!]
    \centering
    \begin{subfigure}[b]{0.49\textwidth}
         \centering
         \includegraphics[width=1\linewidth]{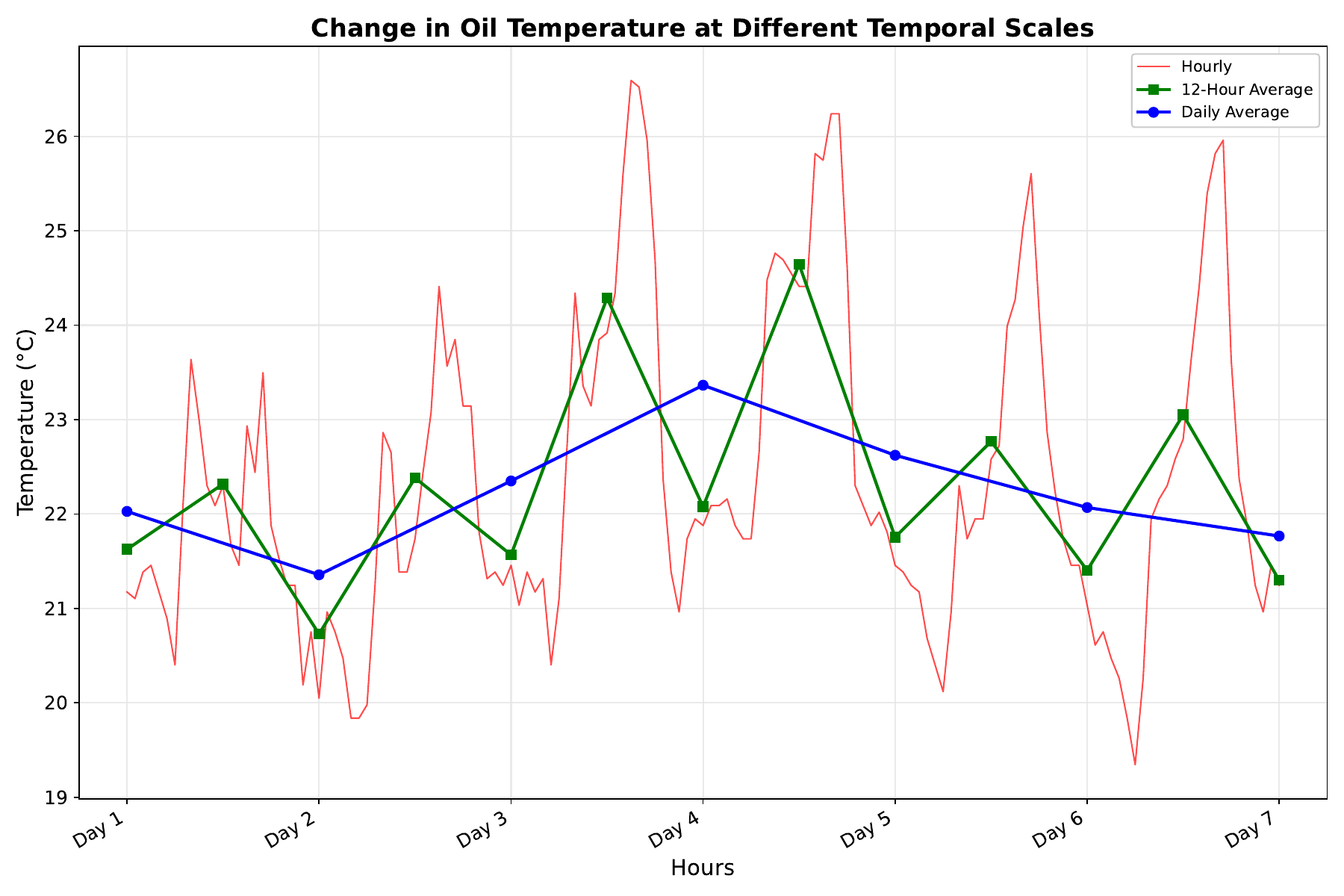}
         \caption{ETTh1 dataset}
         \label{fig:etth1_multiscale}
     \end{subfigure}
    \begin{subfigure}[b]{0.49\textwidth}
         \centering
         \includegraphics[width=1\linewidth]{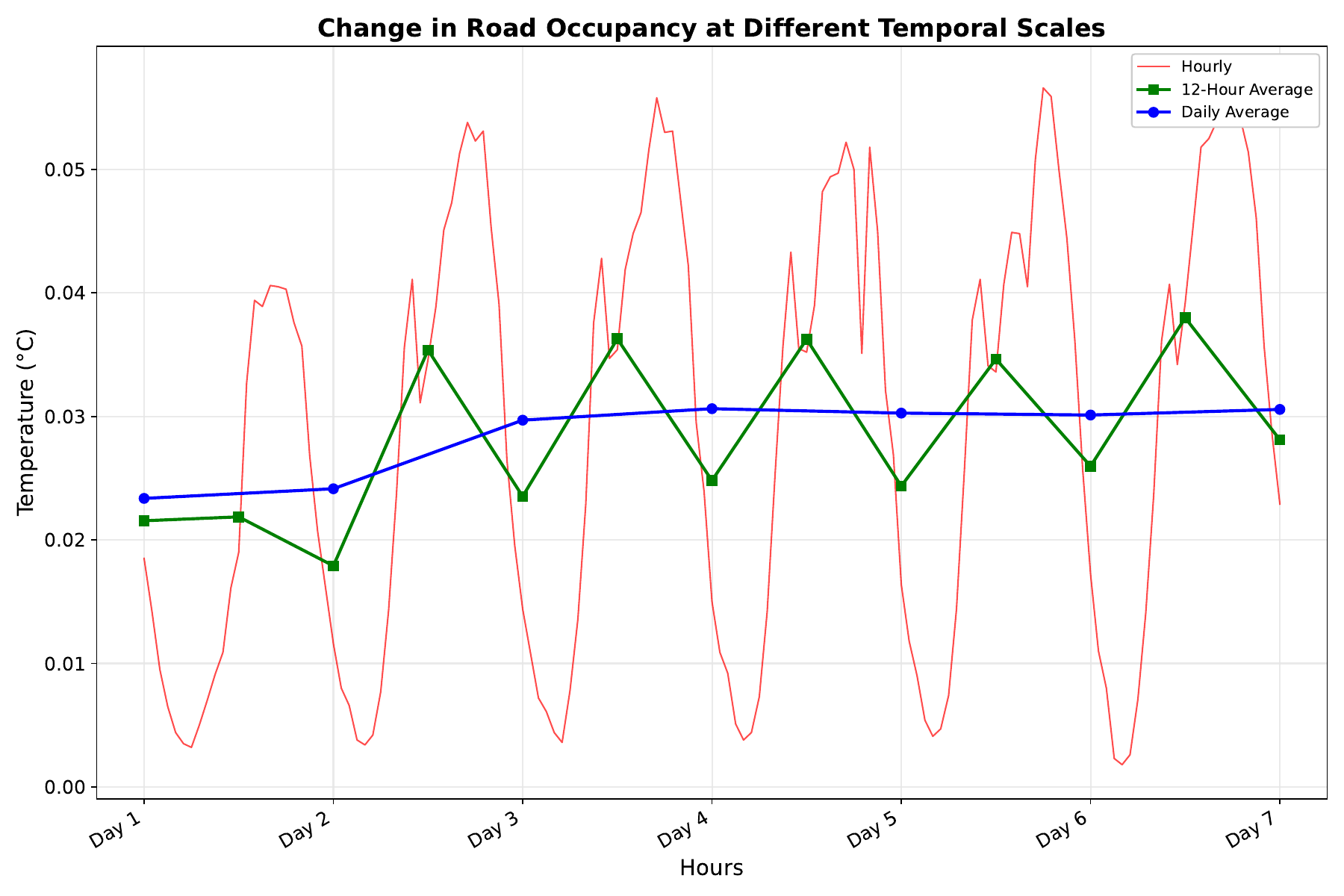}
         \caption{Traffic dataset}
         \label{fig:traffic_multiscale}
     \end{subfigure}
    \caption{\new{Multi-scale visualization for (a) the ETTh1 dataset from ETT category and (b) Traffic dataset. The original signal (red) contains high-frequency fluctuations, while the smoothed versions reveal underlying  and longer-term trends, illustrating the necessity of capturing dynamics at multiple resolutions.}}
    \label{fig:etth2_multiscale}
\end{figure}

\new{To further substantiate the motivation presented in the Introduction, we provide additional visualizations from different domains for one dataset in the ETT category (Figure \ref{fig:etth1_multiscale}) and the Traffic dataset (Figure \ref{fig:traffic_multiscale}). These plots demonstrate that the multi-scale nature of time-series data where signal characteristics change across different temporal resolutions is a ubiquitous phenomenon, not limited to weather data.}
\clearpage

\bibliographystyle{elsarticle-harv} 
\bibliography{references}





\end{document}